# Adapted Foundation Models for Breast MRI Triaging in Contrast-Enhanced and Non-Contrast Enhanced Protocols

Authors:

Tri-Thien Nguyen[1,2], Lorenz A. Kapsner[2,3], Tobias Hepp[4], Shirin Heidarikahkesh[2], Hannes Schreiter[2], Luise Brock[2,5], Dominika Skwierawska[2], Dominique Hadler[2], Julian Hossbach[2], Evelyn Wenkel[6], Sabine Ohlmeyer[2], Frederik B. Laun[2], Andrzej Liebert[2,7], Andreas Maier[1], Michael Uder[2], and Sebastian Bickelhaupt[2]

Affiliations:

[1]Friedrich-Alexander-Universität Erlangen-Nürnberg (FAU), Pattern Recognition Lab, Erlangen, Germany

[2] Institute of Radiology, Uniklinikum Erlangen, Friedrich-Alexander-Universität Erlangen-Nürnberg (FAU), Erlangen, Germany

[3] Friedrich-Alexander-Universität Erlangen-Nürnberg (FAU), Medical Informatics, Erlangen, Germany

[4]Friedrich-Alexander-Universität Erlangen-Nürnberg (FAU), Institute of Medical Informatics, Biometry and Epidemiology, Erlangen, Germany

[5]Friedrich-Alexander-Universität Erlangen-Nürnberg (FAU), Department of Artificial Intelligence in Biomedical Engineering (AIBE), Erlangen, Germany

[6]Radiology München, München, Germany

[7]Institute of Computer Science, Polish Academy of Science, Warsaw, Poland

## ABSTRACT

**Background:** Magnetic resonance imaging (MRI) has high sensitivity for breast cancer detection, but interpretation is time-consuming. Artificial intelligence may aid in pre-screening.

**Purpose:** To evaluate the DINOv2-based Medical Slice Transformer (MST) for ruling out significant findings (Breast Imaging Reporting and Data System [BI-RADS] ≥4) in contrast-enhanced and non–contrast-enhanced abbreviated breast MRI.

**Materials and Methods:** This institutional review board approved retrospective study included 1,847 single-breast MRI examinations (377 BI-RADS ≥4) from an in-house dataset and 924 from an external validation dataset (Duke). Four abbreviated protocols were tested: T1-weighted early subtraction ($T1_{sub}$), diffusion-weighted imaging with b=1500 s/mm² ($DWI_{1500}$), $DWI_{1500}$+T2-weighted (T2w), and $T1_{sub}$+T2w. Performance was assessed at 90%, 95%, and 97.5% sensitivity using five-fold cross-validation and area under the receiver operating characteristic curve (AUC) analysis. AUC differences were compared with the DeLong test. False negatives were characterized, and attention maps of true positives were rated in the external dataset.

**Results:** A total of 1,448 female patients (mean age, 49 ± 12 years) were included. $T1_{sub}$+T2w achieved an AUC of 0.77 ± 0.04; $DWI_{1500}$+T2w, 0.74 ± 0.04 (p=.15). At 97.5% sensitivity, T1sub+T2w had the highest specificity (19% ± 7%), followed by $DWI_{1500}$+T2w (17% ± 11%). Missed lesions had a mean diameter <10 mm at 95% and 97.5% thresholds for both $T1_{sub}$ and $DWI_{1500}$, predominantly non-mass enhancements. External validation yielded an AUC of 0.77, with 88% of attention maps rated good or moderate.

**Conclusion:** At 97.5% sensitivity, the MST framework correctly triaged cases without BI-RADS ≥4, achieving 19% specificity for contrast-enhanced and 17% for non–contrast-enhanced MRI. Further research is warranted before clinical implementation.

## Summary Statement

**A foundation model achieved 19% specificity for contrast-enhanced and 17% for non–contrast-enhanced breast MRI at 97.5% sensitivity for detecting suspicious lesions, supporting further research in triaging.**

## Key Results

- Using a foundation model, T1-weighted early subtraction plus T2-weighted achieved the highest specificity (19%) at 97.5% sensitivity, above the non–contrast-enhanced protocol (17%).

- On the in-house dataset, the model reached an area under the receiver operating characteristic curve of 0.77 with T1-weighted early subtraction plus T2-weighted and 0.74 for the non–contrast-enhanced approach; external validation yielded 0.77.

- At 95% and 97.5% sensitivity, false negatives with T1-weighted early subtraction had mean lesion size <10 mm, mainly non-mass enhancements.

## Abbreviations

| | |
|---|---|
| AI | Artificial Intelligence |
| AUC | Area Under the receiver operating characteristic Curve |
| BI-RADS | Breast Imaging-Reporting and Data System |
| BPD/BPE | Background Parenchymal Diffusion/Enhancement |
| $DWI_{1500}$ | Diffusion-weighted imaging with b-value 1500 |
| MST | Medical Slice Transformer |
| MRI | Magnetic Resonance Imaging |
| NME | Non-Mass Enhancement |
| $T1_{sub}$ | T1-weighted early subtraction |
| T1w/T2w | T1-weighted / T2-weighted |

## Introduction

Breast cancer remains one of the leading causes of cancer mortality in women, with early detection playing a crucial role in treatment outcomes (1). Magnetic resonance imaging (MRI) is a highly sensitive diagnostic tool, particularly for women with dense breast tissue, where mammography is limited (2). However, breast MRI interpretation is time-consuming and requires significant expertise.

The clinical workflow for breast MRI employs the Breast Imaging-Reporting and Data System (BI-RADS) to standardize reporting and management recommendations. A critical decision point in screening is whether a case is most likely benign (BI-RADS 1-3) or requires further immediate diagnostic workup following the screening examination (BI-RADS 4-6). A key challenge in screening workflows is to precisely distinguish between these categories.

Standard breast MRI protocols rely on gadolinium-based contrast agents, which improve diagnostic accuracy but also increase examination time, cost, and carry rare potential health risks (3,4). Consequently, abbreviated MRI protocols, including non-contrast-enhanced approaches such as diffusion-weighted imaging (DWI) using high b-value acquisitions $> 1000$ s/mm$^2$ (5) and virtual contrast-enhanced approaches (6), are being actively investigated to reduce reliance on contrast agents (7).

Artificial intelligence (AI) offers potential for supporting breast MRI analysis, yet the scarcity of large, expertly labeled datasets remains a major barrier for developing AI-based tools in this domain. Pre-trained foundation models address this challenge by leveraging large-scale self-supervised training on diverse datasets, enabling knowledge transfer to medical imaging (8). Despite their potential, foundation models like DINOv2 (9) remain underexplored for 3D medical imaging, particularly in clinically relevant screening contexts such as breast MRI evaluation using BI-RADS classifications.

The Medical Slice Transformer (MST) (10) represents a step toward volumetric self-supervised learning but for breast MRI it has only been validated on the Duke dataset (11) using early contrast-enhanced sequences. This validation, however, presents some limitations: firstly, the dataset contains a disproportionately high number of cancer cases, thus reducing its generalizability to real-world screening populations, and secondly, it does not assess non-contrast sequences.

To further explore foundation models in breast MRI, our study investigates the potential of the MST framework to support triaging based on the presence of suspect lesions warranting immediate diagnostic workup, according to the BI-RADS classification, across four abbreviated breast MRI protocols, including both contrast-enhanced and non-contrast settings.

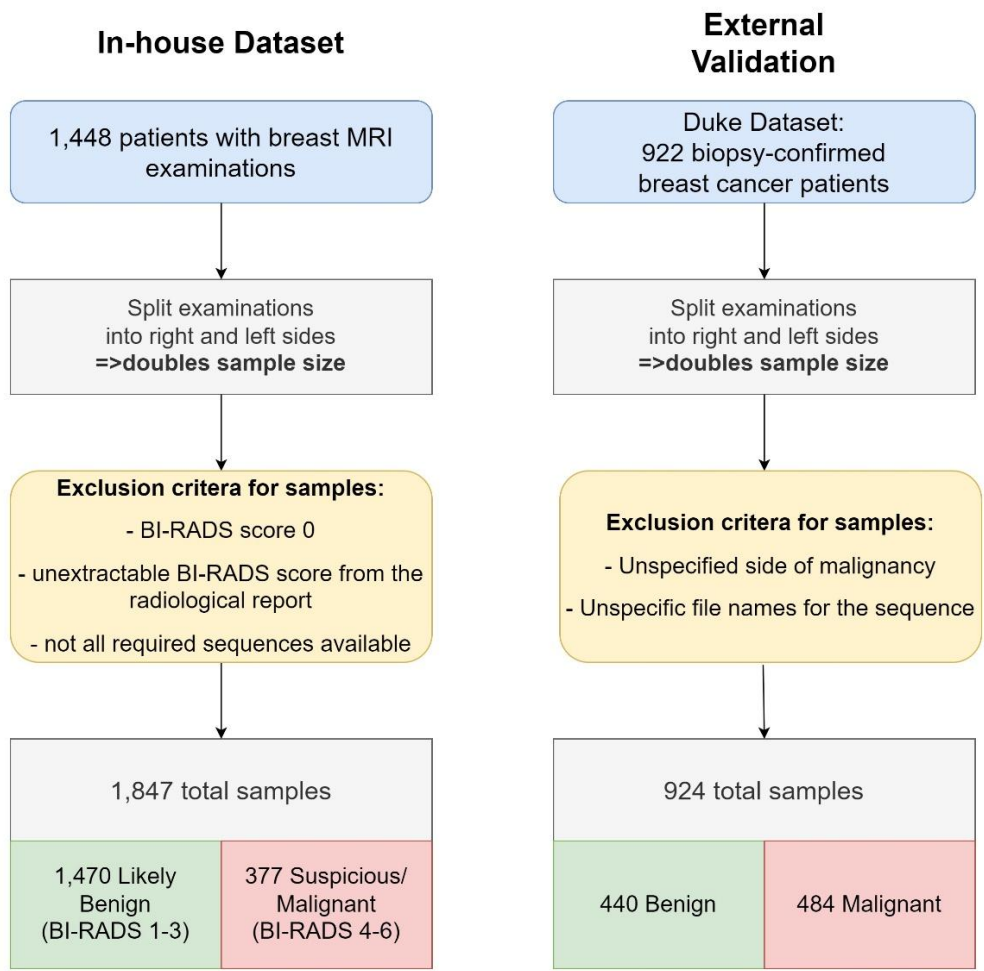

*Figure 1: Dataset generation with exclusion criteria.*

## Materials and Methods

**Study Population and MRI Protocol**

This Institutional Review Board-approved retrospective study (waived informed consent) included n=1,847 single breast MRI examinations and external validation using the publicly available Duke breast MRI dataset (11).

In-house breast MRI examinations:

The cohort comprised 1,448 patients imaged between 2018 and 2022. Examinations were split into left and right breasts; after applying the exclusion criteria shown in Figure 1, 1,847 single breast examinations remained for analysis. Patient characteristics are summarized in Table 1. Approximately 900 examinations (62%) had previously been analyzed in studies focusing on virtual contrast generation (12) and diffusion-weighted artifact detection (13), which addressed different analytic objectives.

MRI was performed on 3-T systems (Vida and Skyra, Siemens) in prone position using 18-channel breast coils. The protocol included (detailed acquisition parameters in Supplemental Table S1):

- Dynamic contrast-enhanced T1-weighted (T1w) 3D gradient echo (TR/TE, 5.97/2.46 ms before and at five time points after gadobutrol injection (0.1 mmol/kg; 60–70 s intervals),

- Axial T2-weighted (T2w) turbo spin echo with fat suppression (TR, 3570–5020 ms; TE, 60–70 ms),

- Axial DWI (TR, 6290–9660 ms; TE, 66–70 ms) with b-values of 50, 750, and 1500 s/mm².

$T1_{sub}$ images were created by subtracting the pre-contrast from the first post-contrast acquisition.

External dataset:

The Duke dataset includes 922 biopsy-confirmed invasive breast cancer cases imaged preoperatively on 1.5-T or 3-T scanners in prone position with contrast-enhanced sequences but no DWI (11).

**Data preprocessing and labeling**

All in-house MRIs were de-identified and transferred to a research workstation. Data included images and radiology reports. BI-RADS scores were extracted from reports using regular expressions and manually verified by a trained medical student. Background parenchymal enhancement (BPE) and diffusion (BPD) were assessed on six axial slices from the $T1_{sub}$ and $DWI_{1500}$ sequences by a board-certified radiologist (T.N., >5 years experience), graded as minimal, mild, moderate, or marked (14).

The Duke dataset already contained binary malignancy labels. Cases with unspecified laterality or unclear naming were excluded, yielding 924 total cases (440 benign, 484 malignant).

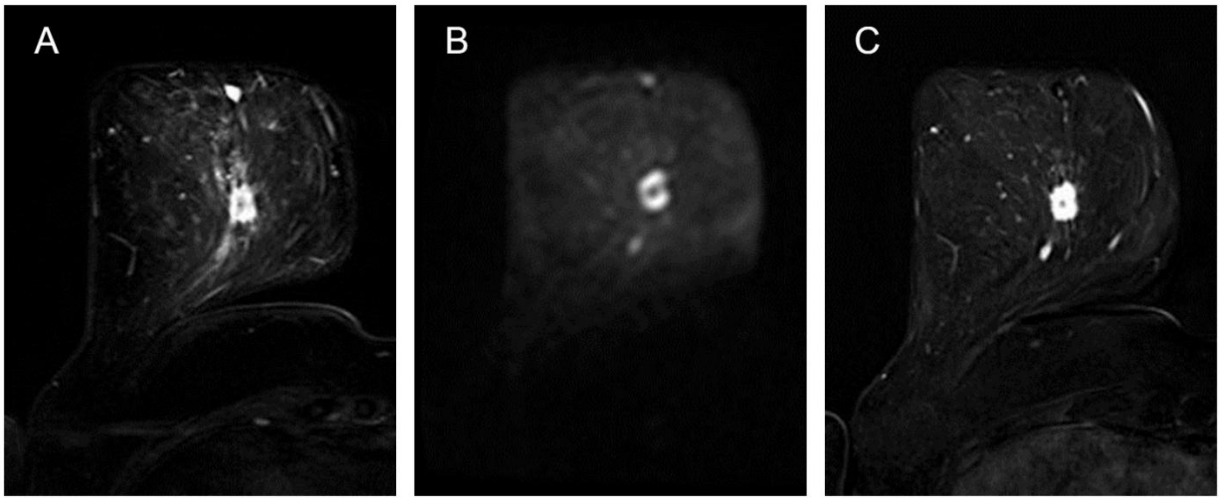

*Figure 2: Sample images from the multiparametric breast MRI protocol showing an axial slice of the three different sequences used in this study. A suspicious lesion is depicted throughout these sequences, shown in T2-weighted image (a), diffusion-weighted image (b), and contrast enhanced T1-weighted subtraction image (c).*

**Image Processing**

All images were preprocessed as described by Liebert et al. (12). The data were split into left and right sides at the x-axis midpoint and resized to 224 × 224 pixels with 38 slices per volume, following Müller-Franzes et al (10) (Figure 2). The following sequence combinations were evaluated for the in-house dataset, while for the external Duke dataset, only $T1_{sub}$ images were available:

- $T1_{sub}$ (both: in-house and Duke dataset)
- $DWI_{1500}$ (in-house)
- T2w with $DWI_{1500}$ (T2w+$DWI_{1500}$) (in-house)
- T2w with $T1_{sub}$ (T2w+$T1_{sub}$) (in-house)

**Network Architecture**

We adapted the MST framework by Müller-Franzes et al. (10) for 3D breast MRI classification. The model uses a pretrained DINOv2 backbone (9) extracts features from 2D slices, which are aggregated by a transformer encoder. A final linear layer performs volume-level classification.

**Model Training**

Data augmentation included random rotations (0–90° around the z-axis), flips, intensity inversions, and Gaussian noise (SD 0.0–0.25). We used 5-fold cross-validation (80/10/10 split). The model was trained with AdamW (lr = 1e-6, cross-entropy loss) on an NVIDIA RTX 6000 GPU.

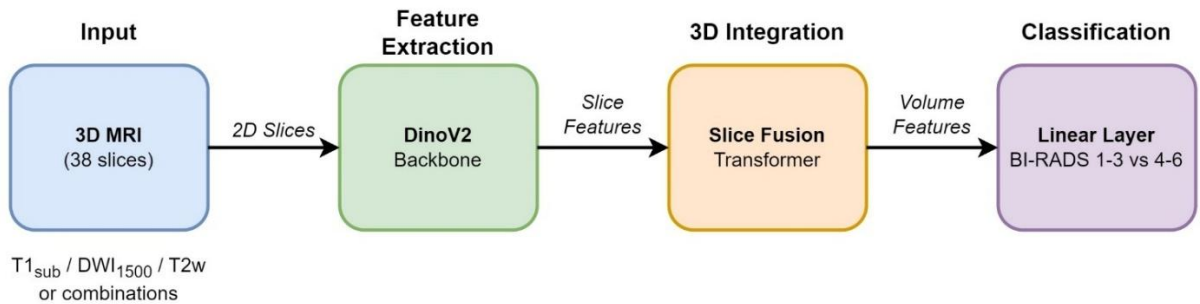

*Figure 3: Medical Slice Transformer architecture for 3D breast MRI classification. $T1_{sub}$, T1-weighted early subtraction; $DWI_{1500}$, diffusion-weighted imaging with b-value 1500 s/mm²; T2w, T2-weighted; BI-RADS, Breast Imaging Reporting and Data System.*

**Evaluation**

For each of the four abbreviated breast MRI protocols, cross-validation folds were evaluated for specificity at high sensitivity thresholds (90%, 95%, 97.5%), relevant to screening applications (15). Area under the receiver operating characteristic curve (AUC) served as an additional metric. False negatives were aggregated across folds and retrospectively reviewed by a board-certified radiologist (T.N., >5 years experience), who categorized lesions as mass, non-mass enhancement (NME), foci (<5 mm), or other (no visible lesion). Attention maps from the MST framework overlaid DINOv2 and transformer attention on input images **(10).**

**External Validation**

For external validation, we tested our $T1_{sub}$ trained model on the Duke dataset to evaluate generalizability Aforementioned radiologist reviewed attention maps generated for true positive cases to assess whether the model correctly identified relevant imaging features using a three point Likert scale (good, moderate, bad).

**Statistical Analysis**

Specificity values at fixed sensitivity thresholds (90%, 95%, and 97.5%) were reported as means with standard deviations across cross-validation folds. Testing across sensitivity thresholds was omitted due to changing decision boundaries.

Pairwise comparisons of the AUC between protocols were performed using DeLong's test **(16)** on fold-level predictions, computed from the combined test sets. The resulting p-values were corrected for multiple testing using the False Discovery Rate method **(17)** . Statistical significance was defined as $p < 0.05$. All analyses were conducted using SciPy (version 1.15.1) All statistical analyses were performed using Python, including SciPy (version 1.15.1) **(18)**, scikit-learn (version 1.6.1) **(19)**, and statsmodels (version 0.14.4) **(20)**.

Code Availability: The source code used for model development and evaluation is publicly available at https://github.com/uker-troi-lab/MultisequenceMST.

*Table 1: In-house patients and sample characteristics*

## Demographics

| Number of patients | 1,448 |
|---|---|
| Gender | Female (100%) |
| Mean age (years) | 49 ± 12 |

## BI-RADS Assessment of Single Breast

| BI-RADS 1 | BI-RADS 2 | BI-RADS 3 | BI-RADS 4 | BI-RADS 5 | BI-RADS 6 |
|---|---|---|---|---|---|
| 22 (1%) | 1,341 (73%) | 107 (6%) | 133 (7%) | 61 (3%) | 183 (10%) |

| Likely benign (BI-RADS 1-3) | 1,470 (80%) |
|---|---|
| Suspicious or malignant (BI-RADS 4-6) | 377 (20%) |

## Parenchymal Assessment of Single Breast

|  | Minimal | Mild | Moderate | Marked |
|---|---|---|---|---|
| Background Parenchymal Diffusion | 376 (20%) | 975 (53%) | 381 (21%) | 115 (6%) |
| Background Parenchymal Enhancement | 511 (28%) | 899 (49%) | 339 (18%) | 98 (5%) |

*Note: BI-RADS = Breast Imaging Reporting and Data System.*

### Results

**In-house study Population Characteristics**

After applying the exclusion criteria, the final analysis included 1,847 single breast MRI examinations with a distribution of 1,470 (80%) likely benign cases (BI-RADS 1-3) and 377 (20%) suspicious or malignant cases (BI-RADS 4-6). BI-RADS 2 was the most common classification, representing approximately 73% cases. The mean age of patients was 49±12 years. For BPD, the majority of cases presented with minimal or mild patterns (73%), similarly to BPE with 76% of examinations as represented in Table 1.

**MST performance on Contrast and Non-contrast Enhanced MRI Protocols**

When training the models, the MST model showed at the 90% sensitivity threshold, the T2w+T1$_{sub}$ combination demonstrated the highest specificity (0.43 ± 0.07, 438/730 false positives aggregated across all cross-validation folds), followed by T1$_{sub}$ alone (0.38 ± 0.14, 490/730 false positives), T2w+DWI$_{1500}$ (0.32 ± 0.05, 504/730 false positives), and DWI$_{1500}$ alone (0.30 ± 0.07, 516/730 false positives).

However, the relative performance ranking partially shifted across thresholds. While the T2w+T1$_{sub}$ combination maintained the highest specificity at 90% and 95% sensitivity (0.43 ± 0.07, 438/730 false positives and 0.29 ± 0.10, 437/577 false positives, respectively), at 97.5% sensitivity, T1$_{sub}$ alone achieved the highest specificity (0.20 ± 0.08, 470/577 false positives). At this highest sensitivity threshold, the remaining combinations showed slightly lower specificity: T2w+T1$_{sub}$ (0.19 ± 0.07, 480/577 false positives), T2w+DWI$_{1500}$ (0.17 ± 0.11, 472/577 false positives), and DWI$_{1500}$ alone (0.14 ± 0.07, 482/577 false positives)

The highest AUC was achieved by T2w+T1$_{sub}$ (0.77 ± 0.04), followed closely by T1sub alone (0.77 ± 0.09). T2w imaging combined with DWI$_{1500}$ achieved an AUC of 0.74 ± 0.04, while DWI$_{1500}$ alone yielded the lowest AUC (0.72 ± 0.04). Overall, no statistically significant differences in AUC were observed between the evaluated sequence combinations. For clarity, *p*-values from pairwise comparisons against the T2w+T1$_{sub}$ reference are reported in Table 2.

Figure 4 presents the receiver operating characteristic curves for all evaluated MRI sequence combinations.

*Table 2: Diagnostic performance metrics of different MRI sequence combinations*

| Sequence | AUC (Mean ± SD) | p-value vs T2w+T1$_{sub}$ | Specificity at 90% Sensitivity (%) | Specificity at 95% Sensitivity (%) | Specificity at 97.5% Sensitivity (%) |
|---|---|---|---|---|---|
| **T1$_{sub}$** | 0.77 ± 0.09 | p=.49 | 37.9 ± 14.0 | 23.5 ± 10.5 | **20.0 ± 8.0** |
| **DWI$_{1500}$** | 0.72 ± 0.04 | p=.21 | 30.5 ± 7.3 | 23.9 ± 8.6 | 14.2 ± 7.2 |
| **T2w+T1$_{sub}$** | **0.77 ± 0.04** | - | **43.3 ± 7.3** | **28.5 ± 10.3** | 19.1 ± 6.9 |
| **T2w+DWI$_{1500}$** | 0.74 ± 0.04 | p=.32 | 32.2 ± 4.6 | 23.9 ± 7.4 | 16.6 ± 10.9 |
| **T1$_{sub}$+T2w+DWI$_{1500}$** | 0.75 ± 0.03 | p=.49 | 32.0 ± 6.7 | 21.4 ± 4.3 | 18.5 ± 3.7 |

*Note: p-values are from DeLong's test for comparing the area under the receiver operating characteristic curve (AUC), calculated on fold-level predictions. Multiple testing was controlled using the false discovery rate method applied to all pairwise sequence comparisons. For brevity, only p-values for comparisons to the T2w+T1$_{sub}$ reference are shown in the table. T1$_{sub}$, T1-weighted early subtraction; T1w, T1-weighted; T2w, T2-weighted; DWI$_{1500}$, diffusion-weighted imaging with b = 1500 s/mm².*

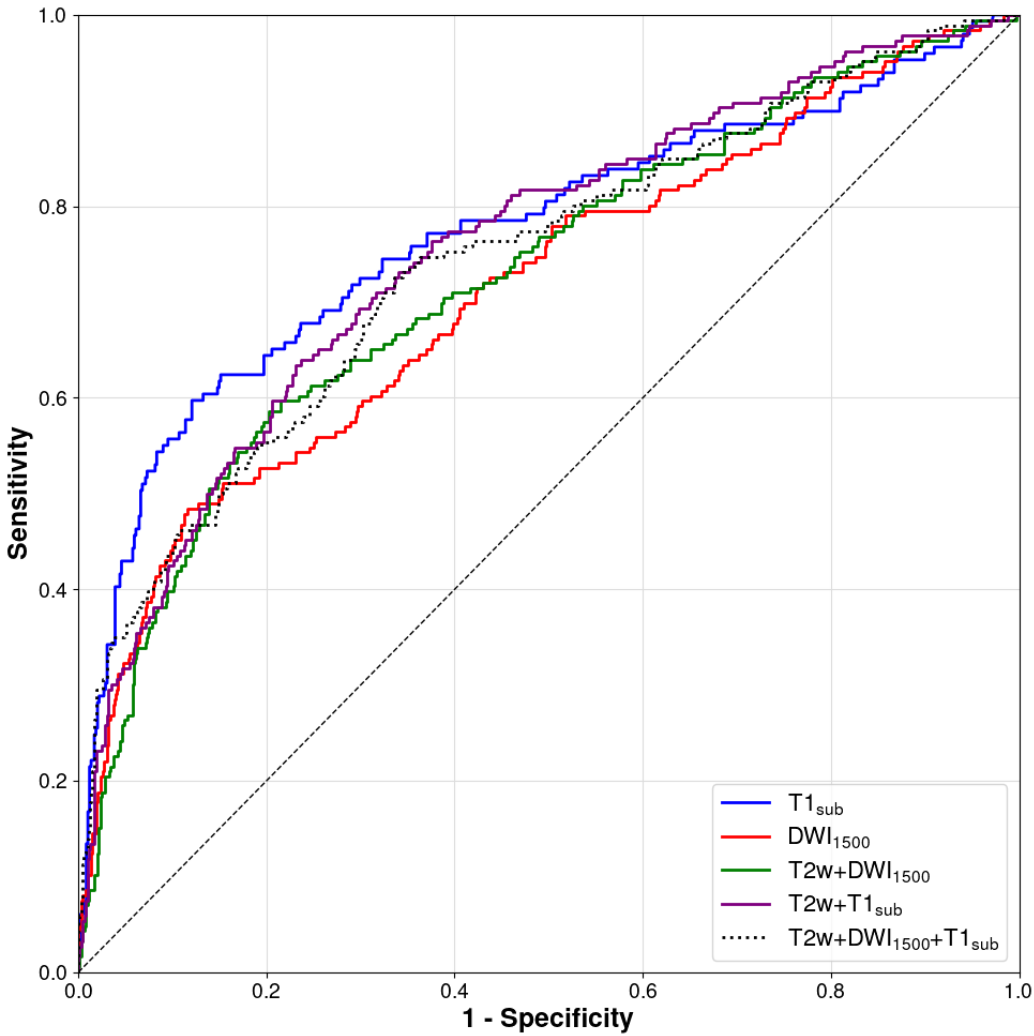

*Figure 4: Receiver operating characteristic curves for the evaluated MRI sequence combinations. The graph shows performance comparisons between different sequence combinations. The diagonal dashed line represents random chance performance. $T1_{sub}$, T1-weighted early subtraction; T1w, T1-weighted; T2w, T2-weighted; $DWI_{1500}$, diffusion-weighted imaging with b = 1500 s/mm².*

**Lesion Characterization of False Negatives**

Analysis of false negative cases pooled across all cross-validation folds at the 90% sensitivity threshold revealed a range of lesion characteristics that were misclassified as likely benign. Table 3 summarizes the characteristics of these missed lesions. Across all sequences, the missed lesions were primarily NME. There was considerable variability in the size of missed lesions, as evidenced by the high standard deviations relative to the mean sizes.

*Table 3: Characteristics of false negatives at different sensitivity thresholds*

| Sequence | Sensitivity threshold | Count | Mean Size (mm) | Mass (%) | NME (%) | Foci (%) | Other (%) |
|---|---|---|---|---|---|---|---|
| $T1_{sub}$ | 90% | 15 | 12±11 | 27 | 33 | 33 | 7 |
| | 95% | 7 | 9±7 | 29 | 29 | 29 | 14 |
| | 97.5% | 3 | 4±5 | 0 | 33 | 33 | 33 |
| $DWI_{1500}$ | 90% | 14 | 14±10 | 25 | 50 | 19 | 6 |
| | 95% | 7 | 6±4 | 29 | 14 | 57 | 0 |
| | 97.5% | 4 | 9±4 | 50 | 25 | 25 | 0 |
| $T2w+DWI_{1500}$ | 90% | 14 | 15±11 | 31 | 44 | 13 | 13 |
| | 95% | 8 | 12±7 | 50 | 25 | 25 | 0 |
| | 97.5% | 4 | 12±9 | 50 | 25 | 25 | 0 |
| $T2w+T1_{sub}$ | 90% | 15 | 11±8 | 18 | 35 | 29 | 18 |
| | 95% | 8 | 12±10 | 0 | 50 | 50 | 0 |
| | 97.5% | 4 | 8±8 | 0 | 25 | 75 | 0 |

*Note: Non-mass enhancement (NME). "Other" denotes cases in which no lesion was visible and the diagnostic category was determined by other means (eg, histopathologically confirmed lesion). $T1_{sub}$, T1-weighted early subtraction; T1w, T1-weighted; T2w, T2-weighted; $DWI_{1500}$, diffusion-weighted imaging with b = 1500 s/mm².*

**External Validation Metrics**

For external validation, the model was trained on the full in-house dataset using the T1sub sequence and evaluated on the Duke dataset. This yielded an AUC of 0.77. At the operating point of 90% sensitivity, the specificity was 43 % as shown in Figure 5. The model demonstrated promising generalization capabilities, though performance metrics were lower compared to cross-validation results on the primary dataset.

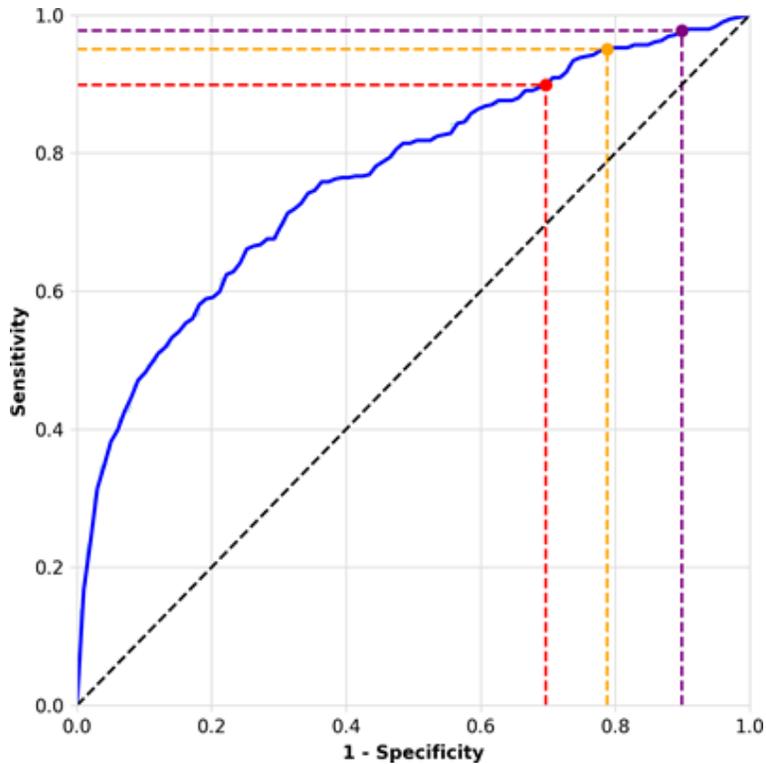

*Figure 5: Performance of the T1-weighted early subtraction model on the Duke dataset. Receiver operating characteristic curve with horizontal dashed lines marking sensitivity thresholds: red, 90%; yellow, 95%; and purple, 97.5%.*

**External Validation Explainability**

Attention maps were generated and evaluated for all 226 true positive cases for $T1_{sub}$ from the Duke dataset. Radiological evaluation of the attention maps showed predominantly favourable results. For area attention maps, 53% were rated as good and 35% as moderate, with only 12% receiving a bad rating. Similarly, slice attention maps showed strong performance with 57% rated as good and 31% as moderate, with just 12% rated as bad. Overall, 88% of both attention map types were considered acceptable (good or moderate) as shown in Table 4.

Figure 6 shows exemplary images of both attention map types, demonstrating how the model identifies relevant regions and slices containing lesions.

*Table 4: Radiologist ratings of attention maps for 226 true positive cases.*

| Rating | Area attention (n, %) | Slice attention (n, %) |
| --- | --- | --- |
| **good** | 119 (53) | 129 (57) |
| **moderate** | 80 (35) | 70 (31) |
| **bad** | 27 (12) | 27 (12) |

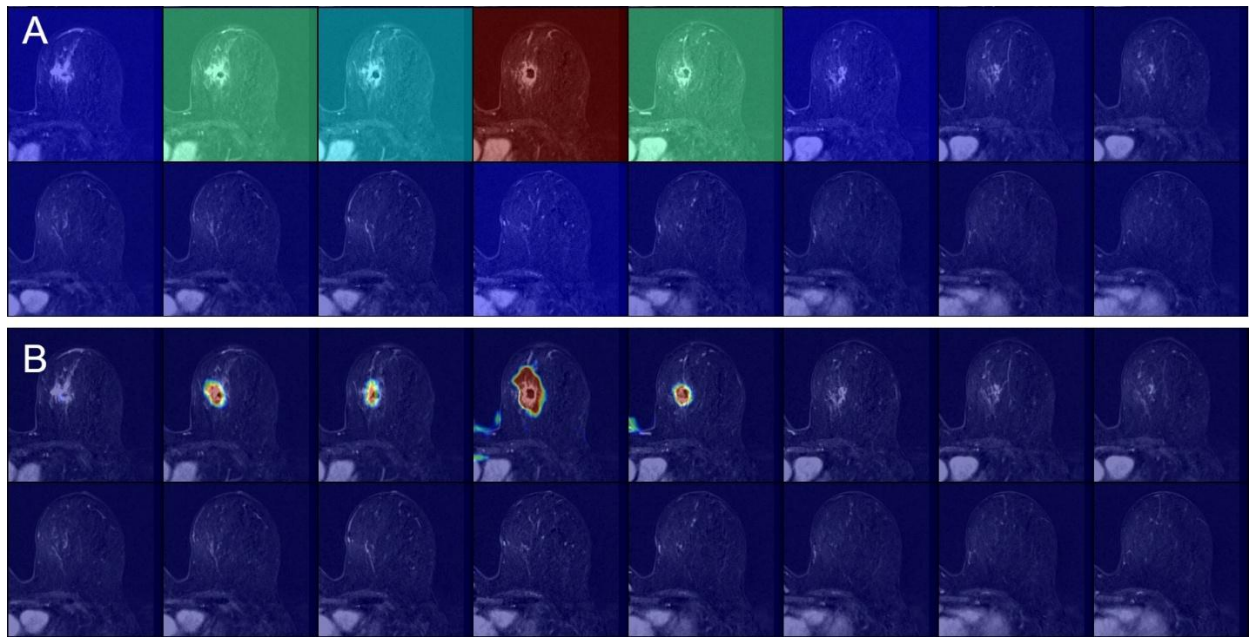

*Figure 6: Attention maps for a true positive case. (A) Slice attention highlighting the most important slices, and (B) Area attention displayed as a heat map showing regions of interest within slices.*

## Discussion

Our study evaluated using MST for triaging breast MRI examinations to rule out BI-RADS 4-6 lesions. The approach was tested on abbreviated contrast-enhanced and non-contrast enhanced MRI protocols, with the latter being based on high b-value diffusion weighted imaging at b=1500s/mm2 ($DWI_{1500}$). For the input combination T2w+$T1_{sub}$, the highest AUC was achieved (0.77 ± 0.04), with corresponding specificities of 28% and 19% at 90% and 95% sensitivity, respectively. Interestingly, the non–contrast-enhanced combination T2w+$DWI_{1500}$ showed no statistically significant difference in AUC compared to the contrast-enhanced reference, although it achieved slightly lower specificities of 24% and 17% at the same sensitivity thresholds.

Traditional mammography screening is limited in women with high breast density (21). Supplemental MRI screening might benefit these women (22,23) and has proven valuable for high-risk patients, improving early detection and diagnostic accuracy (24). This has led to its incorporation into screening guidelines for specific risk groups (25), however widespread implementation of breast MRI screening remains limited despite evidence supporting its cost-effectiveness for women with dense breast tissue (26,27).

Challenges such as limited radiologist availability and the time-intensive demands of MRI interpretation might contribute to this gap. AI tools could help address these barriers by improving efficiency and reducing costs. For conventional mammography, AI screening has been successfully tested in a retrospective study (28). With AI as a second reader, large randomized trials have been performed with comparable results to human readers alone (29).

In their overview, Youk and Kim (30) highlight that research in AI applications for screening breast MRI is still limited. Verburg et al. (31) used a simpler classification scheme (BI-RADS 1 vs. 2–5) and 2D projections, achieving a higher AUC (0.83), while our more clinically relevant 3D classification (BI-RADS 1–3 vs. 4–6) yielded 0.77.

Abbreviated MRI protocols are increasingly investigated (7). We evaluated MST on different abbreviated sequence combinations, including non-contrast and contrast-enhanced protocols. Non-contrast protocols showed no statistically significant diagnostic performance loss compared to contrast-enhanced protocols, despite slightly lower specificity. This supports further evaluation of unenhanced breast MRI with high b-value DWI for screening as previously investigated (32,33), which may particularly benefit from AI integration.

The overall performance level observed from the MST model however still remained lower than human readers in comparable screening scenarios. Kuhl et al. (34) demonstrated that radiologists achieve specificities of up to 94% even when using abbreviated protocols. This substantial performance gap indicates that current AI models, while promising, remain complementary tools rather than standalone replacements for expert radiologists.

Analysis of the missed lesions revealed that they also included lesions >10 mm at lower sensitivity thresholds. At the 95% and 97.5% sensitivity thresholds, the mean size of non-detected lesions was below 10 mm for both $T1_{sub}$ and $DWI_{1500}$; however, with the addition of T2w imaging, the average size of missed lesions slightly exceeded 10 mm. The good performance on smaller lesions may be attributable to the use of 3 T state-of-the-art MRI systems and a breast-optimized DWI protocol with multiple averages (n = 15/20) for the b = 1500 s/mm² acquisition, highlighting the diagnostic potential of optimized breast DWI.

NME constituted the most frequently missed group, suggesting that the network may struggle to capture some of the variable aspects of malignant lesions. However, examination of the attention maps for true positive cases demonstrated that in 88% of instances, the attention maps showed good or moderate overlap of the lesion, supporting the explainability of the classification.

External validation on the Duke dataset using our $T1_{sub}$-trained model yielded an AUC of 0.77, compared to 0.94 achieved by Müller-Franzes et al. (10). This difference is likely attributable to variations in classification criteria, as our study used BI-RADS scores (1–3 vs. 4–6) rather than a benign versus histologically proven malignant categorization.

This study has several limitations. We used a limited single-center dataset and lacked prospective validation. Our binary classification approach doesn't fully leverage BI-RADS granularity, focusing on clinical thresholding rather than histopathological stratification. We investigated only high b-value DWI, not lower b-values common in clinical practice. External validation involved domain shift between our BI-RADS training approach and Duke dataset's histopathological stratification. We did not compare MST to other AI approaches. Future research should explore alternative architectures, multimodal approaches, and clinical data integration across different clinical settings to determine broader applicability.

In conclusion, this study explored the potential of using an adapted foundation model for triaging breast MRI examinations in a clinical setting. The narrow performance gap between contrast-enhanced protocols and diffusion-weighted images, coupled with benefits from additional sequences, suggests viable pathways for optimizing breast MRI screening approaches that can improve early detection and outcomes.